\pdfoutput=1

\documentclass[11pt]{article}

\usepackage{EMNLP2022}

\usepackage{times}
\usepackage{latexsym}

\usepackage[T1]{fontenc}

\usepackage[utf8]{inputenc}

\usepackage{microtype}

\usepackage{inconsolata}

\usepackage{amsthm}
\usepackage{mathtools}
\usepackage{graphicx}
\usepackage{wrapfig}
\usepackage{multirow}
\usepackage{subcaption}
\usepackage{algorithm}
\usepackage{algorithmic}
\usepackage{booktabs}
\usepackage{amssymb}

\title{Explainable Slot Type Attentions to Improve \\ Joint Intent Detection and Slot Filling}

\author{Kalpa Gunaratna\textsuperscript{$\dagger$}, Vijay Srinivasan\textsuperscript{$\dagger$}, Akhila Yerukola\textsuperscript{$\ddagger$}\thanks{\: Work done while at Samsung Research America}, \and Hongxia Jin\textsuperscript{$\dagger$} \\
        \textsuperscript{$\dagger$}{Samsung Research America, Mountain View CA, USA}\\
        \texttt{\{k.gunaratna, v.srinivasan, hongxia.jin\}@samsung.com}\\
        \textsuperscript{$\ddagger$}{Carnegie Mellon University, Pittsburgh PA, USA}\\
        \texttt{ayerukol@andrew.cmu.edu}}

\begin{document}
\maketitle
\begin{abstract}
Joint intent detection and slot filling is a key research topic in natural language understanding (NLU). Existing joint intent and slot filling systems analyze and compute features collectively for all slot types, and importantly, have no way to explain the slot filling model decisions. In this work, we propose a novel approach that: (i) learns to generate additional \textit{slot type specific features} in order to improve accuracy and (ii) provides \textit{explanations} for slot filling decisions for the first time in a joint NLU model. We perform an additional constrained supervision using a set of binary classifiers for the slot type specific feature learning, thus ensuring appropriate attention weights are learned in the process to explain slot filling decisions for utterances. Our model is \textit{inherently explainable} and does not need any post-hoc processing. We evaluate our approach on two widely used datasets and show accuracy improvements. Moreover, a detailed analysis is also provided for the exclusive slot explainability.
\end{abstract}

\section{Introduction}
\begin{figure}[]
    \centering
    \includegraphics[scale=0.39, trim=0cm 14.5cm 6cm 0cm, clip=true]{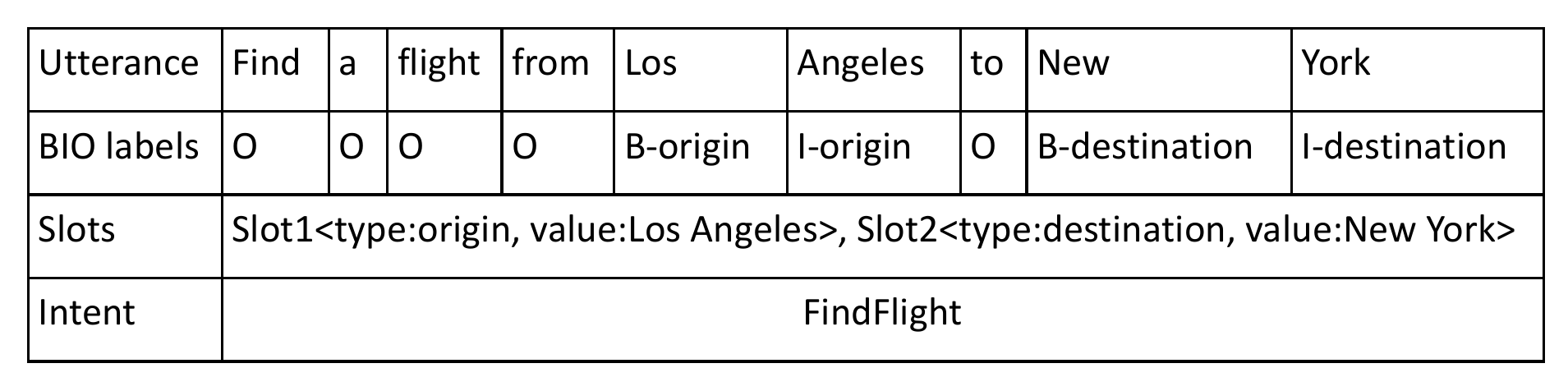}
    \caption{Intent detection and slot filling example.}
    \label{fig_intro}
\end{figure}

Natural language understanding (NLU) is a critical component in building intelligent interactive agents such as Amazon Alexa, Google Assistant, Microsoft's Cortana, and Samsung's Bixby. It helps to understand the intricate details of user utterances, including: (i) understanding the intents of the utterances that help determine the agent's actions and (ii) detecting slots that signify important entities and phrases required to complete the actions. Figure~\ref{fig_intro} shows an example utterance with an intent and two slots (a slot is a type and value pair). Typically, slot classification (i.e., slot filling) is viewed as a sequence labeling problem which is handled through BIO sequence labeling notation as shown in Figure~\ref{fig_intro}.

State-of-the-art techniques in NLU jointly optimize both the intent detection and slot filling tasks~\cite{chen2019bert}, including various attention, gating, and transformer-based architectures~\cite{liu2016attention, qin2021co}. However, existing techniques treat all slot types collectively. For example, they may compute a general attention weights vector to represent all the slot types. Therefore, they do not learn slot type specific patterns, and features based on those patterns that can potentially increase the model performance. In this work, we show that both learning to capture slot type specific patterns through attention weights and computing new features from them `explicitly' is beneficial in at least two ways: (i) boosting joint NLU model \textit{accuracy}, and (ii) making the model \textit{explain} its slot filling decisions at a granular level. To properly train our slot type specific attention weights learning, we introduce an auxiliary network of binary classifiers to the joint model that has an additional optimization objective. Auxiliary networks and respective optimization objectives in NLP are often used to enforce additional constraints ~\cite{joshi2020spanbert, zhao2020learning}. We use an auxiliary network to enforce slot type specific attention learning.

On the other hand, explainable AI has gained attention in NLP~\cite{danilevsky2020survey} as it provides `explanations' for `black box' deep learning model decisions. The terms `interpretable AI' and `explainable AI' have been used interchangeably in the literature~\cite{danilevsky2020survey}. To be consistent and avoid confusion~\footnote{Some literature consider \textit{interpretability} to be harder and requires achieving human cognition/performance/agreement.}, we use the term `explainable' in this work to refer to a model's ability to provide explanations to the outside world of its inner working, similar to ~\cite{wiegreffe2019attention}. According to ~\cite{jacovi2020towards}, explainable NLP models fall into two main categories: (i) models that require external post-hoc processing for explainability, and (ii) models that are inherently explainable due to the capability built into the model. The latter is argued to be preferred in the literature~\cite{rudin2019stop}. Explainable models are transparent~\cite{lipton2016mythos} and hence, can gain trust. However, existing joint NLU approaches disregard this important explainability aspect. Our proposed joint NLU model in this work is inherently explainable and explains slot filling decisions for each slot type, which is at a granular level. To the best of our knowledge, our approach is the first to inherently support fine-grained explainability for slot filling in a joint NLU model. Our contributions in this work are two-fold:
\begin{enumerate}

    \item \textbf{Accuracy Improvement:} We model per slot type fine-grained attentions and compute new features in addition to intent and slot features to \textit{improve} accuracy of joint intent detection and slot filling task. This is enforced through an auxiliary network of binary classifiers constrained to the number of slot types and supervised using automatically generated data.

    \item \textbf{Slot Explainability:} We show that per slot type fine-grained attention weights computed through the supervision of the auxiliary network can be used to \textit{explain} the slot filling task in the joint NLU model.

\end{enumerate}

We evaluate our approach using ATIS and SNIPS benchmark datasets and show improvements. We also provide a detailed analysis on the versatility and robustness of slot type specific attention weights in providing insights into slot filling.

\section{Related Work}
Intent detection and slot filling tasks are two typical sub-tasks of NLU. Slot filling is generally challenging as decisions have to be made for each word/token and can be seen applied in interesting use cases~\cite{gunaratna2021using}. The two tasks were performed independently in the past~\cite{haffner2003optimizing,raymond2007generative}. 
However, jointly optimizing them has shown better accuracy recently (e.g., ~\cite{goo2018slot,chen2019bert,qin2019stack,hakkani2016multi,qin2021co}). Moreover, recent advances in contextual language models have improved the language encoding capabilities that are directly reflected in joint NLU models~\cite{chen2019bert,qin2021co} compared to traditional static word embedding approaches~\cite{hakkani2016multi}. Existing joint NLU models leverage RNN~\cite{liu2016attention,goo2018slot}, CNN~\cite{xu2013convolutional}, intent influenced attention~\cite{dao2021intent}, gated attention~\cite{goo2018slot}, self attention~\cite{li2018self,zhang2020graph,he2020syntactic}, stack propagation~\cite{qin2019stack}, hierarchical information flow~\cite{lee2018coupled,zhang2019joint,zhang2020graph}, multi task learning~\cite{pentyala2019multi}, memory networks~\cite{liu2019cm}, syntactic information integration~\cite{wang2021encoding}, and custom transformer architectures~\cite{qin2021co}. 

The models mentioned above can learn correlations and hierarchy that exist between intents and slots, but compute features collectively over all slot types. Therefore, these correlations or attention mechanisms are not useful to explain slot filling decisions because (i) they are abstract and do not correspond to each individual slot filling decision, and (ii) may focus on general language semantics or special tokens ~\cite{clark2019does}. Also, most deep neural models are `black boxes' and hence, not transparent and explainable \cite{guidotti2018survey}, making them hard to earn trust and fix errors. Hence, explainable systems have gained attention for NLP problems such as sentiment classification and question answering~\cite{danilevsky2020survey} and outside NLP~\cite{rai2020explainable,zhang2018explainable}. Importantly, Rudin argues that developing inherently explainable models is important because use of post-hoc techniques to explain existing systems leads to many problems~\cite{rudin2019stop}. Ours is the first inherently explainable method for slot filling in a joint NLU model.

In our proposed model, we compute slot type specific attention weights to explain slot filling decisions. There has been a debate on whether attentions are explainable~\cite{jain2019attention,wiegreffe2019attention}. However, it has been shown that attentions can be used for explanations, if they contribute towards the model outcomes/behaviors~\cite{wiegreffe2019attention}. We show that this is true in our model and provide evidence for a strong correlation between the slot type attentions and slot predictions (Section~\ref{sec:interpretability}). Orthogonal to ours, having multiple final slot classifiers is possible to model separate features but it introduces the big challenge of resolving overlapping spans (i.e., nested NER)~\cite{wang2020pyramid}.

\section{Approach}

\begin{figure*}
    \centering
    \includegraphics[scale=0.74, trim=0cm 9.2cm 1.55cm 0.25cm, clip=true]{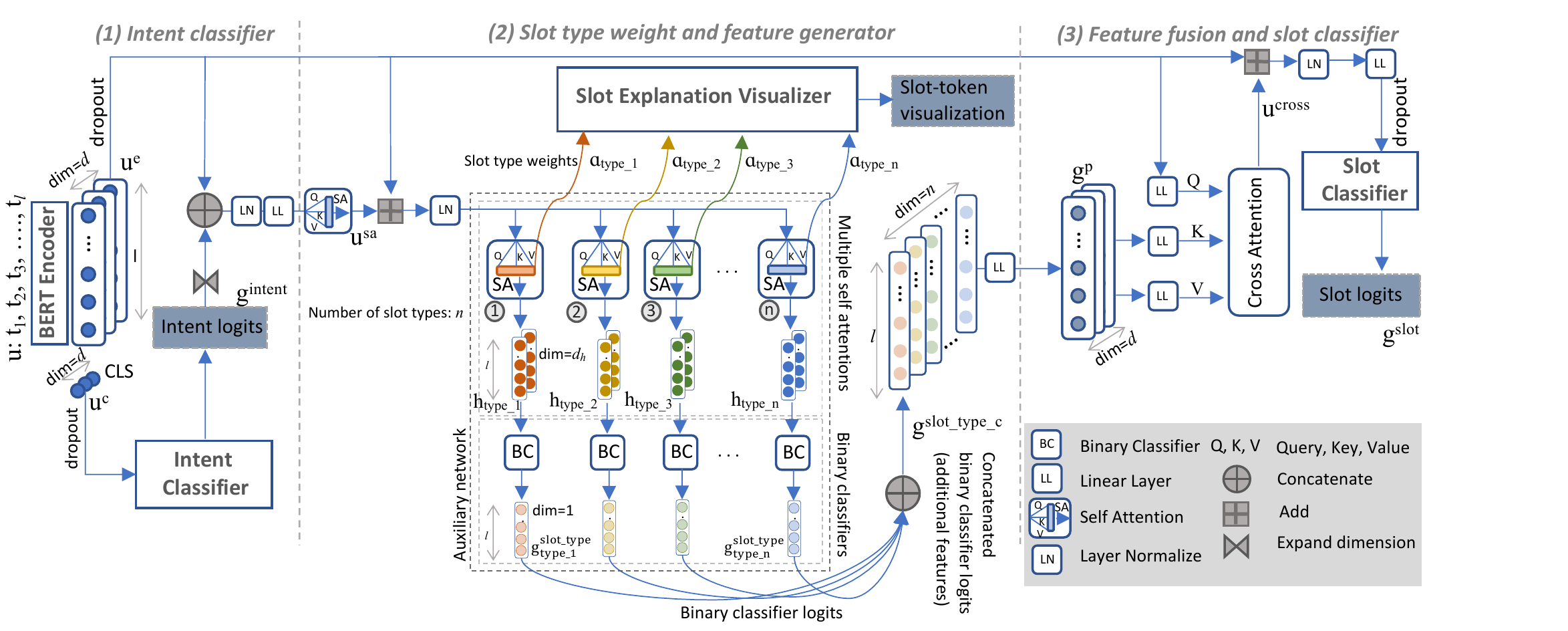}
    \caption{Overview of our explainable joint NLU model. $n$ is the total number of slot types, $l$ is the utterance length, $d_h$ is the slot type converted utterance embedding dimension, and $d$ is the original utterance embedding dimension. In addition to intent and slot predictions, our model visualizes per slot type attentions as explanations. We model per slot type features and learn appropriate weights - different colors show each slot type modeling.}
    \label{fig_architecture}
    \vskip -0.1in
\end{figure*}

Our approach consists of three network components: (i) intent classifier, (ii) slot classifier, and (iii) slot type weight and feature generator. Existing joint NLU models consist of the first two network components arranged in various configurations whereas ours is distinguishably different from them because of the third auxiliary network component. This third component learns to compute: (i) separate per slot type features (i.e., logits) and (ii) fine-grained per slot type attention weights for each utterance token. Further, it is supervised through automatically generated training data (see Section~\ref{sec:slot_type_classifier}) and constrained to exactly the number of slot types. Our per slot type attention weights can provide insights on the focus points of the input utterance that are relevant to labeling a particular slot type, which is absent in other approaches. Our approach is illustrated in detail in Figure~\ref{fig_architecture}.

\subsection{Network}

Let $l$ be the number of tokens in the input utterance $u$=[$t_1$, $t_2$, $t_3$, ..., $t_l$] where $t_i$ represents the $i$th token in $u$. Let $T$ be the original slot type set in the ground truth (e.g., location, time, etc.), $S$ be the set of sequence labeling format (BIO) labels (e.g., O, B-location, I-location, etc.), and $I$ be the set of intents in the ground truth (e.g., FindFlight, Search, etc.). It is important to model dynamics of O (in BIO) in our model and hence, O $\in T$.

Given an utterance $u$, we first tokenize and encode it using a contextual language model like BERT. Then we get the encoded utterance $u^e$=$[\boldsymbol{t_1}, \boldsymbol{t_2}, \boldsymbol{t_3}, ..., \boldsymbol{t_l}]$ ($\in \mathbb{R}^{l \times d}$) where token $\boldsymbol{t_i}$ represents the embedding vector for the $i$th token, $\boldsymbol{t_i} \in \mathbb{R}^d$ and $d$ is the embedding dimension.

\subsubsection{Intent Classifier}
Intent classifier is a single layer multi-class classifier that uses BERT as the language encoder. For a given utterance $u$, the embedding vector for [CLS] token is the context embedding $u^c \in \mathbb{R}^{d}$. Intent classifier outputs intent logits $g^{intent}$ as follows.
\begin{equation}
    g^{intent} = u^c W^{intent} + b^{intent}
    \label{eq_intent_logits}
\end{equation}

where, $W^{intent} \in \mathbb{R}^{d \times |I|}$ is the learnable weight matrix and $b^{intent}$ is the bias.
$Y^{intent}$ is the one hot encoded intent vector and $p^{intent}=softmax(g^{intent})$. Intent loss $\mathcal{L}^{intent}$ for utterance $u$ is computed using cross entropy loss as follows.

\begin{equation}
    \mathcal{L}^{intent} = - \sum_{x}^{|I|} Y_x^{intent} log(p_x^{intent})
\end{equation}

\subsubsection{Slot Type Weight and Feature Generator}
\label{sec:slot_type_classifier}
We discuss details of our auxiliary network that helps the model learn explainable weights and generate general yet slot type specific features. 
First, after applying dropout to the encoded utterance $u^e$, we concatenate intent logits $g^{intent}$ with $u^e$. Note that $g^{intent} \in \mathbb{R}^{|I|}$, hence we expand the logits vector by one dimension, copy the values along the expanded dimension $l$ times (utterance length) to get the expanded intent logits tensor $g^{int\_e}$ ($\in \mathbb{R}^{l \times |I|}$). Concatenation of intent logits with token embeddings and then applying self attention provides direct fusion of intent features with the slot classification network to learn any correlations between intents and slots. We compute the new feature $u^{sa}$ ($\in \mathbb{R}^{l \times d}$) as follows. $LL$, $LN$, and $SA$ represent linear layer, layer normalization, and self attention functions, respectively. $\oplus$ is the concatenation. 

\begin{equation}
\label{eq:u-sa}
    u^{sa} = SA(LL(LN(u^e \oplus g^{int\_e}; \theta^{LN}); \theta^{LL}); \theta^{SA})
\end{equation}

$\theta^{LN}$, $\theta^{LL}$, and $\theta^{SA}$ are the sets of model parameters. Self attention function $SA(.)$ is as follows.
\begin{equation}
    SA(x; \theta^{SA}) = softmax \left(\frac{Q_x K_x^{T}}{\sqrt{d_x}} \right) V_x
    \label{eq_self_attention}
\end{equation}

Query $Q_x$, key $K_x$, and value $V_x$ are calculated from input $x$ using three different linear projections, $LL_q(x)$, $LL_k(x)$, and $LL_v(x)$, respectively. $d_x$ is the dimension of the input $x$ (also $K_x$). In Equation~\ref{eq:u-sa}, $LL$ projects back to dimension $d$ and $d_x$ = $d$ in $SA$.
Then we add the original utterance embedding $u^e$ to $u^{sa}$ and perform a layer normalization to get $\hat{u^e}$ before the auxiliary network. 

\begin{equation}
    \hat{u^e} = LN(u^e + u^{sa}; \theta)
\end{equation}

\subsubsection*{Slot type constrained self attention weights} 

We project the utterance embedding $\hat{u^e}$  into multiple representations; exactly $|T|$ times. We want our model to explain slots at a fine-grained level and hence, we need weight distributions over the entire input for each token, for each slot type. We compute multiple self attentions, one per slot type, that provides slot type specific attention weights per token with respect to the utterance. 

However, without proper supervision, these attention weights are not  meaningful. Therefore, we make these multiple projections correspond to $|T|$ slot type binary classifiers. 
We apply the self attention defined in Equation~\ref{eq_self_attention}, $|T|$ number of times, one for each slot type by projecting different query $Q_{type\_i}$, key $K_{type\_i}$, and value $V_{type\_i}$ tensors ($\in \mathbb{R}^{l \times d_h}$) from $\hat{u^e}$, each using three different linear projections. Self attended output $h_{type\_i}$ ($\in \mathbb{R}^{l \times d_h}$) and attention weights $\alpha_{type\_i}$ ($\in \mathbb{R}^{l \times l}$) for slot type $type\_i$ ($\in T$) are as follows. $d_h$ is the embedding dimension.

\begin{equation}
\begin{aligned}
    h_{type\_i} 
    = softmax \left( \frac{Q_{type\_i} K_{type\_i}^{T}}{\sqrt{d_h}} \right) V_{type\_i} \\
    \alpha_{type\_i} = softmax \left( \frac{Q_{type\_i} K_{type\_i}^{T}}{\sqrt{d_h}} \right)
\end{aligned}
\end{equation}

\subsubsection*{Slot type feature generation and supervision}

\begin{figure}[]
    \centering
    \includegraphics[scale=0.35, trim=0.6cm 11.8cm 3.6cm 0cm, clip=true]{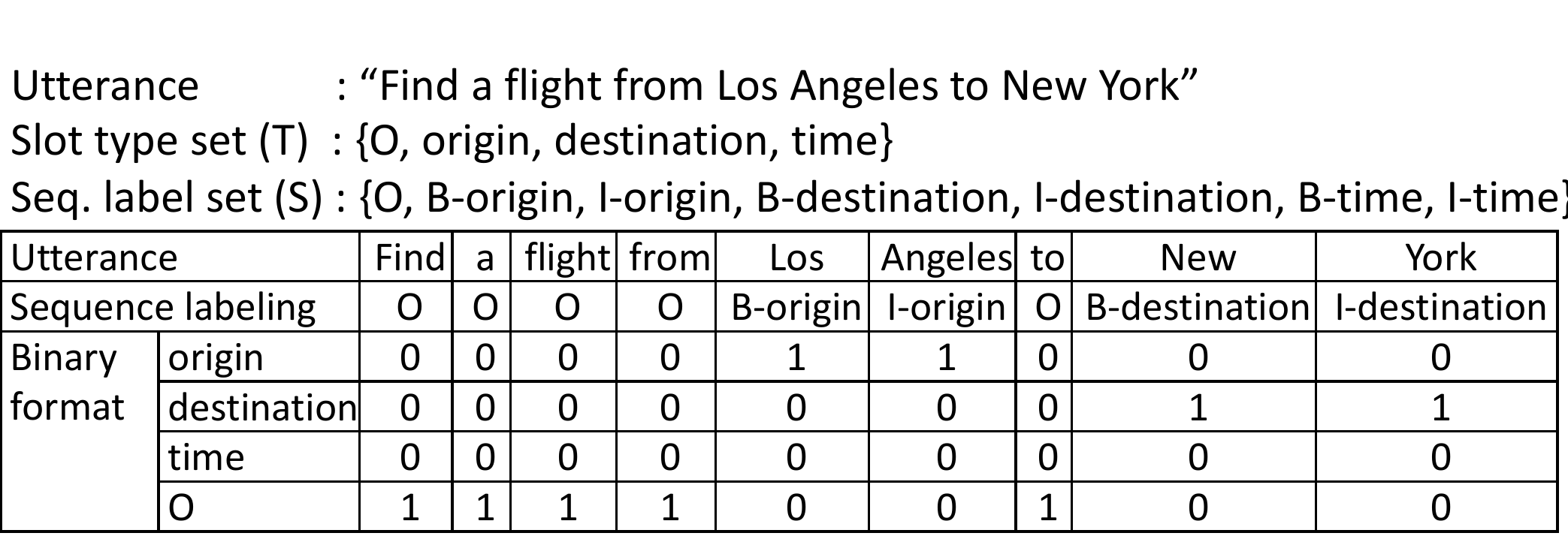}
    \caption{Data generation for the auxiliary network.}
    \label{fig_preprocessing}
\end{figure}

Slot type projections require additional optimization objectives so that the self attention weights are truly meaningful and not random noise. We train each slot type projection to predict binary output that states whether an input token in the utterance is true (1) or false (0) for that slot type. This training data is automatically generated from sequence labeling BIO ground truth as shown in the example in Figure~\ref{fig_preprocessing}. Binary format for a slot type except for O is generated by having 1s to slot tokens (non-O positions) and 0s to otherwise. For O, all non-slot tokens are marked as 1s and rest are 0s.

With the additional optimization objectives for each slot type, self attention weights can be used to visualize attention points of the input for each token with respect to each slot type as explanations. This is because: (i) per slot type, the embedding for the token being classified is computed attending to all the input tokens including itself and (ii) the same token embedding is used for the particular slot type classification (controlled supervision), and further, (iii) type classification output logits are fed to the final slot classification as additional features (explicit connection to final classification). Our binary classification output logits enable the model to learn general patterns/features specific to each slot type that also boost model performance.
Binary classification for $i$th slot type $type\_i$ ($\in T$) is performed as follows. We initialize a set of weight tensors $W^H$ and bias $b^H$ where $|W^H|$ = $|b^H|$ = $|T|$. 

\begin{equation}
    g^{slot\_type}_{type\_i} = h_{type\_i} W^H_{type\_i} + b^H_{type\_i}
\end{equation}
where $h_{type\_i} \in \mathbb{R}^{l \times d_h}$, $W^H_{type\_i} \in \mathbb{R}^{d_h \times 1}$, $b^H_{type\_i} \in \mathbb{R}$, and slot type logits $g^{slot\_type}_{type\_i} \in \mathbb{R}^{l \times 1}$. Binary cross entropy loss $\mathcal{L}^{type}$ for type classification for utterance $u$ is as follows.

\begin{small}
\begin{equation}
    \mathcal{L}^{type} = - \frac{1}{N} \left( \sum_{j=1}^{N} Y_j^{t} log(p_j^{t}) + (1-Y_j^{t})log(1-p_j^{t}) \right)
\end{equation}
\end{small}

where, $Y^{t}$ is the one-hot encoded ground truth vector and $p^t$ is the predicted probabilities vector, and $N$ is the total number of data points. Note that we measure binary cross entropy loss collective for all the slot type classifiers. Therefore, if $|u| = l$, then $N = l \times |T|$.

\subsubsection*{Slot explanation visualizer} 
We take the attention weights matrix $\alpha_{type\_i}$ for a slot type $type\_i$ and visualize attentions on the utterance for each token. See Figure~\ref{fig_visualize} for examples.

\subsubsection{Feature Fusion and Slot Classifier}

Slot type logits generated from the previous step are new features that go into our slot classifier to improve accuracy. They are slot type specific and hence, can capture fine-grained slot patterns. We concatenate all the slot type logits per each input utterance token and then apply cross attention on it by having query $Q_{u^e}$ projected from the utterance $u^e$ and key $K_{g^p}$ and value $V_{g^p}$ projected from the concatenated and projected slot type logits $g^p$. The logits concatenation is performed at the token level where all the slot type logits for a token are concatenated to produce the tensor ${g}^{slot\_type\_c} (\in \mathbb{R}^{l \times |T|}$). Then the concatenated logits tensor is projected to get $g^p$ ($\in \mathbb{R}^{l \times d}$). We then apply cross attention between utterance embedding $u^e$ and $g^p$ to get slot type aware utterance representation $u^{cross}$ ($\in \mathbb{R}^{l \times d}$). We compute query, key, and value tensor projections ($\in \mathbb{R}^{l \times d}$) as follows. $\theta^1$, $\theta^2$, and $\theta^3$ are layer parameters for the three different linear projection layers.

\begin{math}
Q_{u^e}=LL(u^e;\theta^1), K_{g^p}=LL(g^p;\theta^2), V_{g^p}=LL(g^p;\theta^3)
\end{math}

\begin{equation}
    u^{cross} = softmax \left(\frac{Q_{u^e} K_{g^p}^{T}}{\sqrt{d}} \right) V_{g^p}
\end{equation}

Cross attention between the utterance embeddings and slot type logits highlights slot type specific features from the utterance embeddings. This is added to utterance as follows to make the slot classifier input $u^{slot}$ ($\in \mathbb{R}^{l \times d}$).

\begin{equation}
    u^{slot} = LL(LN(u^e + u^{cross}; \theta^{s_{LN}}); \theta^{s_{LL}})
\end{equation}

Finally, the slot logits tensor $g^{slot}$ ($\in \mathbb{R}^{l \times |S|}$) is computed where $W^{slot} \in \mathbb{R}^{d \times |S|}$.

\begin{equation}
    g^{slot} = u^{slot} W^{slot} + b^{slot}
\end{equation}

The slot loss $\mathcal{L}^{slot}$ for utterance $u$ is computed using the cross entropy loss. $Y_{i,s}^{slot}$ is the one hot encoded ground truth slot label and $p_{i,s}^{slot}$ is the softmax probability of slot BIO label $s$ for $i^{th}$ token.

\begin{equation}
    \mathcal{L}^{slot} = - \sum_{i}^{l} \sum_{s}^{|S|} Y_{i,s}^{slot} log(p_{i,s}^{slot})
\end{equation}

\subsubsection*{Network optimization}
We optimize our network by minimizing the joint loss of the three sub-networks. Overall loss $\mathcal{L}$ for utterance $u$ is defined as in Equation~\ref{eq:total_loss}. $\alpha$, $\beta$, and $\gamma$ are hyperparameters that represent loss weights.

\begin{equation}
    \mathcal{L} = \alpha \mathcal{L}^{intent} + \beta \mathcal{L}^{type} + \gamma \mathcal{L}^{slot}
    \label{eq:total_loss}
\end{equation}

\section{Evaluation and Discussion}

\begin{table*}[]
\centering
\begin{tabular}{lllll}
\hline
\multirow{2}{*}{Model}                          & \multicolumn{2}{c}{SNIPS}    & \multicolumn{2}{c}{ATIS}      \\ \cline{2-5} 
                                                & Intent & Slot & Intent & Slot \\ \hline
RNN-LSTM~\cite{hakkani2016multi}                & 96.9              & 87.3      &  92.6             & 94.3                  \\
Attention-BiRNN~\cite{liu2016attention}         & 96.7              & 87.8      &  91.1             & 94.2                  \\
Slot-Gated~\cite{goo2018slot}                   & 97.0              & 88.8      &  94.1             & 95.2                  \\ \hline
Joint BERT~\cite{chen2019bert}                  & 98.6              & 97.0      &  97.5             & 96.1                  \\
Joint BERT + CRF~\cite{chen2019bert}            & 98.4              & 96.7      &  97.9             & 96.0                  \\
Co-Interactive Transformer~\cite{qin2021co}        & 98.8              & 95.9      &  97.7             & 95.9                  \\
Co-Interactive Transformer + BERT~\cite{qin2021co} & 98.8              & 97.1      &  98.0             & 96.1                  \\ \hline
Ours (BERT as the language encoder)                 & \textbf{98.99}     & \textbf{97.24} &  \textbf{99.10}& \textbf{96.20}                  \\
\hline
\end{tabular}
\caption{Joint intent detection and slot filling results on the two benchmark datasets. Our novel slot type specific attention model for joint NLU task improves the state-of-the-art model performance.
}
\label{tab:eval_accuracy}
\end{table*}

\begin{table*}[]
\centering
\begin{tabular}{lllll}
\hline
\multirow{2}{*}{Model}                          & \multicolumn{2}{c}{SNIPS}    & \multicolumn{2}{c}{ATIS}      \\ \cline{2-5} 
                                                & Intent     & Slot         & Intent     & Slot \\ \hline
Our Full Model                                            & \textbf{98.99}        & \textbf{97.24}    & \textbf{99.10}        & \textbf{96.20} \\
(-) slot type feature and weight generator                           & 98.85                 & 96.77             & 98.54                 & 96.00          \\
(-) cross attention         & 98.71                 & 96.57             & 99.10                 & 95.76         \\
(-) intent logit concatenation with utterance                    & 98.99                 & 96.44             & 99.10                 & 95.87         \\
\hline
\end{tabular}
\caption{Ablation experiments of our approach. (-) refers to removing that part from the network in Figure~\ref{fig_architecture}.}
\label{tab:eval_ablation}
\end{table*}

We use two most widely used datasets for joint NLU: ATIS~\cite{hemphill1990atis} and SNIPS~\cite{coucke2018snips}. ATIS has 4478, 500, and 893 utterances for train, dev, and test splits, respectively. SNIPS has 13084, 700, and 700 utterances for train, dev, and test splits, respectively. We measure slot accuracy on the exact span match and maximum utterance length is 50.

\paragraph{Hyperparameters} We experimented with batch size, and number of epochs and set learning rate and dropout as in JointBert~\cite{chen2019bert}(appendix ~\ref{sec:appendix_parameters}). We experimented with 20, 30, and 40 epochs, batch sizes of 32, 64, and 128, and trained the model for the entirety of the epochs\footnote{We saved the model at the end of all training epochs. Training used only the training dataset split; dev data split was used for hyperparameter selection.}. Using 40 and 20 epochs with batch size 32 gave the best performance on ATIS and SNIPS datasets, respectively. For all the experiments, we used learning rate of 5e-5, dropout of 0.1, and Adam optimizer.
In our experiments, we set $\alpha$=$\beta$=$\gamma$=1 and the slot type network attention (projection) dimension $d_h$=32. We implemented our approach in PyTorch and used BERT-Base (any encoder works). We use BertViz~\cite{vig2019multiscale} for visualizations.

\subsection{Joint NLU Model Accuracy}

We consider several comparable baselines to evaluate against our approach. RNN-LSTM~\cite{hakkani2016multi} uses LSTMs to predict the intent and slot labels whereas Attention-BiRNN~\cite{liu2016attention} is an extension which uses attention mechanisms. Slot-Gated~\cite{goo2018slot} introduced gating mechanism on LSTMs to improve joint intent detection and slot filling. These approaches represent the pre-contextual language model era. The advent of contextual language models such as BERT have enabled significant improvements in NLU models. Joint BERT~\cite{chen2019bert} uses BERT as the language encoder, using the [CLS] token for intent classification and token embeddings for slot filling. Co-Interactive Transformer~\cite{qin2021co} uses modified transformer layers after the initial BERT encoding. We also use BERT as the language encoder.

As shown in Table~\ref{tab:eval_accuracy}, our joint NLU approach with the addition of slot type attentions and auxiliary network yields the best performance. The slot type specific modeling not only improves the slot filling performance but also the intent detection accuracy. We also see in Table~\ref{tab:eval_accuracy} that BERT and transformer-based NLU models~\cite{chen2019bert,qin2021co} outperform sequence labeling models based on RNN or LSTM. Our model shows accuracy improvements that go even further, due to its novel slot type specific feature computation and fusion. 

We also present an ablation study by removing 3 components from Figure~\ref{fig_architecture} as shown in Table~\ref{tab:eval_ablation}. We see that there are significant drops in both the intent and slot task performance when individual components are removed. Additionally, it is interesting to see that dropping the cross attention with the slot type logit features and utterance embedding affects the performance more than dropping the entire slot type auxiliary network. This may be because computing features is not sufficient; appropriately fusing them is necessary and important.

\subsection{Slot Explainability}
\label{sec:interpretability}

\begin{figure*}[ht!]
    \centering
    \includegraphics[scale=0.73, trim=0cm 11.5cm 3cm 0cm, clip=true]{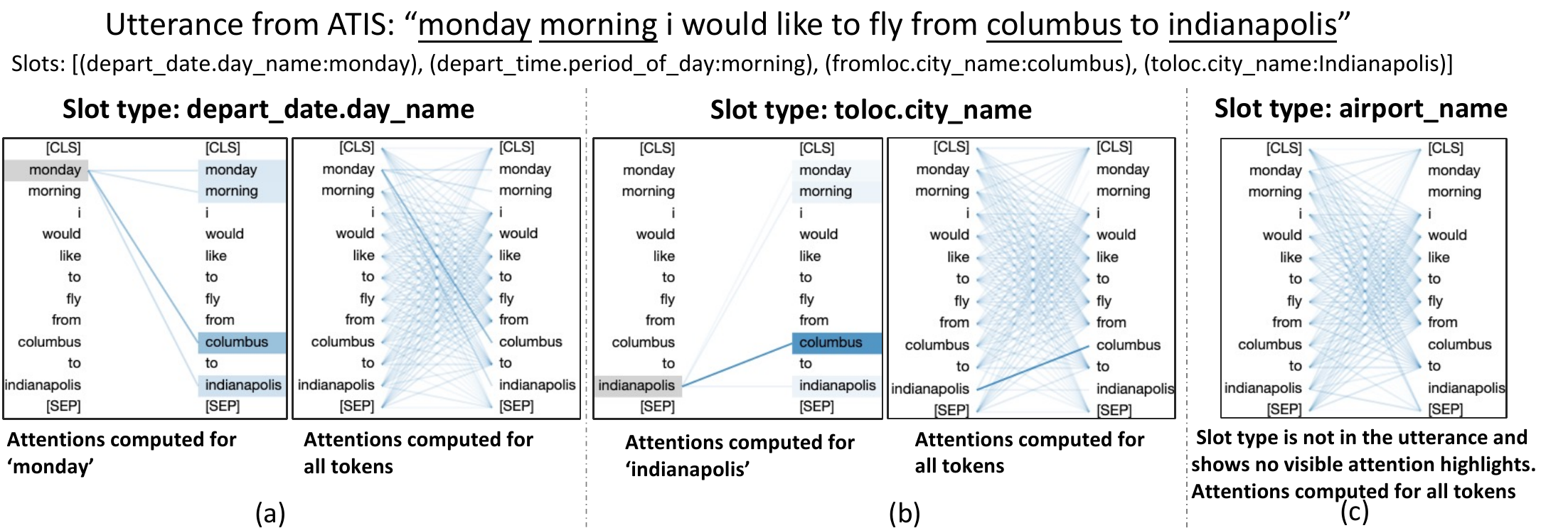}
    \caption{Slot type specific attentions and explainable visualizations on an example utterance from ATIS. (a) and (b) show plots for two `positive' slot types (that appear in the utterance) whereas (c) shows a plot for a `negative' slot type. Blue rectangles in the left plots in (a) and (b) show the attention points of the model for the tokens in the gray rectangle and the target slot type.}
    \label{fig_visualize}
\end{figure*}

A novel feature of our joint NLU model is slot explainability. Using attentions as explanations has caused some debate in the NLP community~\cite{jain2019attention,wiegreffe2019attention}. However, ~\cite{wiegreffe2019attention}~\footnote{They further stated "Attention mechanisms do provide a look into the inner workings of a model, as they produce an easily-understandable weighting of hidden states."} argued that one can use attentions to provide explanations if they are `necessary for good performance', and suggested several experiments for this purpose. We present our findings along this line in Table~\ref{tab:frozen_attn} where model training and predictions were performed while freezing slot type attention weights to be a uniform distribution. We can see that then the model performance drops, which implies that these attention weights represent important and meaningful features.

\begin{table}[]
\centering
\small
\begin{tabular}{llll}
\hline
\multicolumn{2}{c}{SNIPS} & \multicolumn{2}{c}{ATIS} \\
Intent        & Slot       & Intent       & Slot   \\ \hline
98.99         & 96.88      & 98.99        & 95.78     \\ \hline
\end{tabular}
\caption{Accuracy while freezing slot type attentions (from a uniform distribution).}
\label{tab:frozen_attn}
\end{table}

Slot type specific attention weights reflect where the model focuses on the utterances. Figure~\ref{fig_visualize} shows computed attention weights belonging to some slot types when visualized against an example utterance taken from ATIS. Slot types \textit{depart\_date.day\_name} and \textit{toloc.city\_name} are two positive slot types (out of 4) for the utterance. In Figure~\ref{fig_visualize} (a) and (b), we can see where the model is concentrating for the particular slot type. This is only possible because we model per slot type feature computation through self attention (note that simple query-based attentions will not work for explainability), supervision (via auxiliary network), and also fusion in the main network. 
For example, consider attention weights for \textit{toloc.city\_name} slot type, to classify `indianapolis' as \textit{toloc.city\_name}, model pays attention to token `columbus' which belongs to \textit{fromloc.city\_name}. Similarly for the token `monday' that belongs to type \textit{depart\_date.day\_name}, the model pays attention to \textit{fromloc.city\_name} and \textit{toloc.city\_name} positions.
In Figure~\ref{fig_visualize}(c), we show attention weights computed for the slot type \textit{airport\_name} that is not present in the utterance (i.e., negative slot type) and all the attentions computed for it are equally low, with no significant attention points. 
It is evident from this example that using slot type attentions, we are able to see the general patterns that the model looks at for each slot type.

\begin{figure}[]
    \centering
    \includegraphics[scale=0.48, trim=0cm 12.8cm 8.8cm 0cm, clip=true]{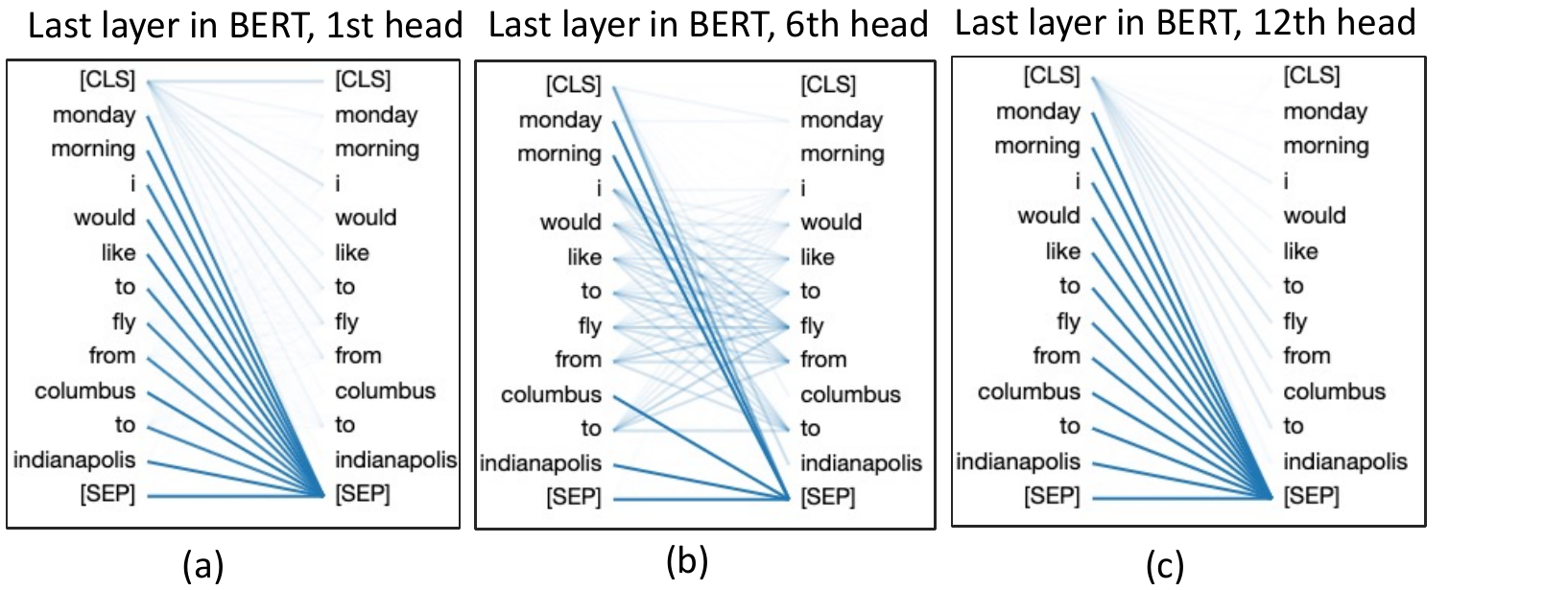}
    \caption{Attentions in the last BERT layer. (a), (b), and (c) shows first, sixth, and twelfth attention heads.}
    \label{fig_visualize_bert}
\end{figure}

\begin{figure}[]
    \centering
    \includegraphics[scale=0.37, trim=0.53cm 1.2cm 1.5cm 0cm, clip=true]{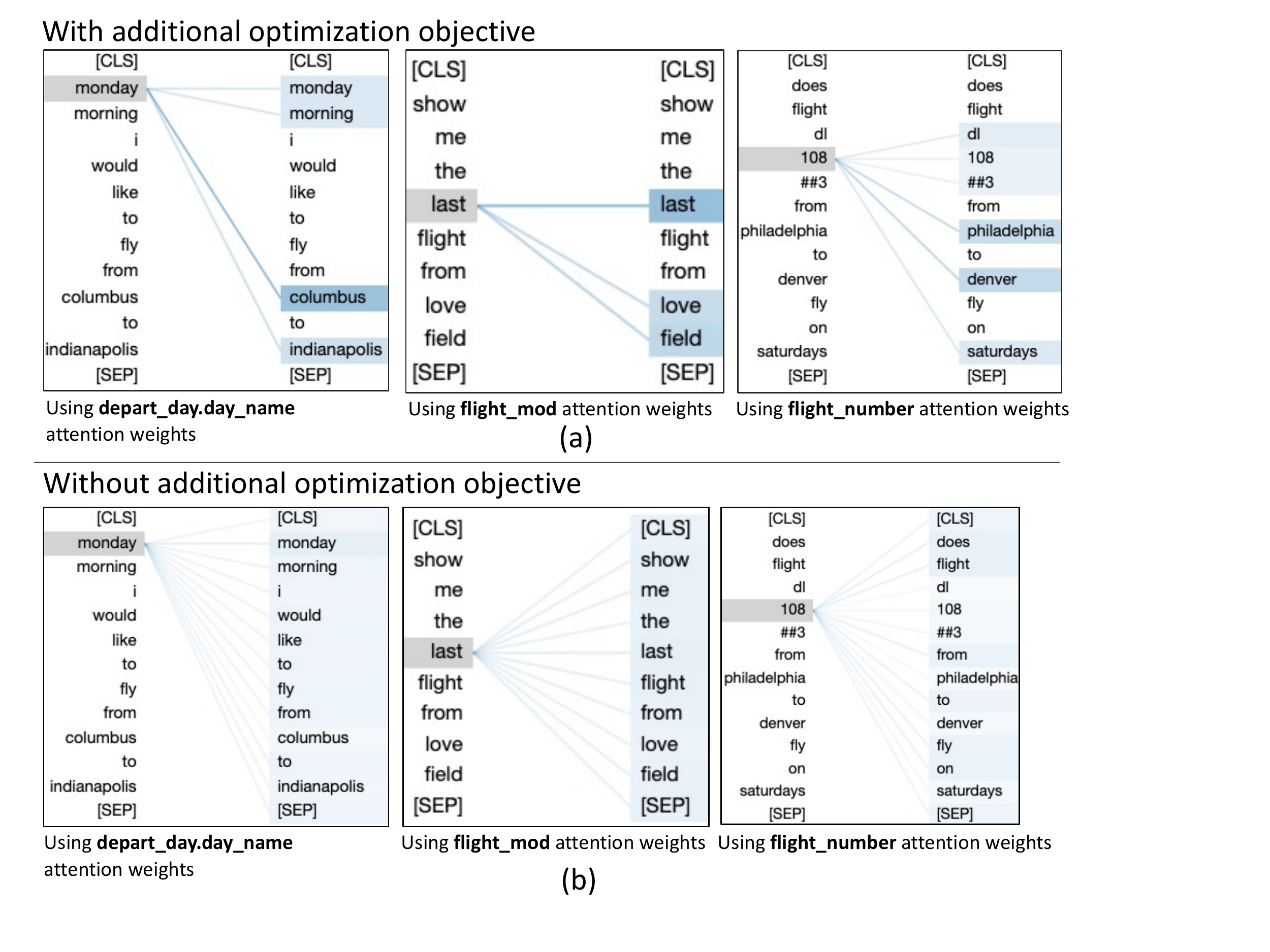}
    \caption{Slot type specific attention weights have to be properly supervised. In these visualizations, (a) shows that our auxiliary network design enables proper attention computations whereas, (b) shows that simply performing multiple attention computations and fusing does not yield proper attentions.}
    \label{fig_visualize_test}
\end{figure}

In fact, explicit modeling and network design is required to provide explanations like we have shown. For example, consider multi-head attention layers in BERT (multiple heads in the 12 layers in BERT base). Despite multiple heads being deployed to learn and compute multiple feature vectors, they do not correlate with the slot types as there is no 1-1 mapping of slot types to attention heads through any supervision; hence, these attention heads cannot be used for slot explanations. Clark et al.~\cite{clark2019does} analyzed BERT layer attentions and they observed that some attentions mainly focus on general language semantics and special tokens (e.g., [SEP]). In Figure~\ref{fig_visualize_bert}, we show several attention heads in the last BERT layer, and we see that they cannot be used to explain slot labeling decisions for each slot type, this is also true for any attention attribution techniques (e.g., \cite{hao2021self}), self attention or traditional attention methods applied on the utterance since a single attention weight vector cannot be broken down to multiple slot types. 

To further illustrate this effect, we removed the auxiliary binary classification (i.e., slot type output optimization) that uses the automatically generated data in the training phase and used the resulting model to visualize attentions for the slot types. See Figure~\ref{fig_visualize_test}(a), where it shows that the network learns proper attention weights for each slot type with our auxiliary network compared to Figure~\ref{fig_visualize_test}(b), where it shows when the additional optimization is removed, network is not guaranteed to learn any meaningful attention weights vector.

\subsubsection{Quantitative Analysis}

It is important to verify that the observations shown in Figure~\ref{fig_visualize} are consistent. Our hypothesis is that positive slot type attention weights (Figure~\ref{fig_visualize} (a) and (b)) have clear spikes or specific focus areas in attention compared to the negative slot type attention weights (Figure~\ref{fig_visualize} (c)). We use entropy\footnote{\url{https://en.wikipedia.org/wiki/Entropy_(information_theory)}} to capture this difference. We expect positive slot type attention weights to have lower entropy compared to the negative slot type attention weights, which tend to have a more uniform distribution, hence, higher entropy. We compute entropy for a list of attention weights [$x_1$, $x_2$,.., $x_n$] using the equation below, where, $Px_i$ = $x_i$/$\sum x_i$.

\begin{equation}
    entropy = - \sum Px_i \times log_2(Px_i)
\label{eq:entropy}
\end{equation}

We compute the average entropy of the positive slot types and the negative slot types separately for each utterance and then average the entropy values. As shown in Table~\ref{tab:entropy_list}, the positive slot types achieve lower average entropy values compared to the negative slot types for both the datasets. The table also shows a breakdown of top $k\%$ attention weights (sorted in descending order and take the first $k\%$ items). We can see that when $k$ is small, like 5\% or 10\%, the entropy difference between positive and negative slot types is higher. The attention vectors are long, and hence several attention spikes for positive slot types may not seem prominent in the entropy computation when the entirety of the attention weights is considered. These results suggest that there are more attention spikes in positive slot types for utterances compared to negative slot types. Thus, we can use the distinguishable attention spikes to gain insights into slot filling decisions.

\begin{table}[]
\footnotesize
\setlength\tabcolsep{2.5pt}
\begin{tabular}{r|lll|lll}
\hline
\multicolumn{1}{c|}{\multirow{2}{*}{\begin{tabular}[c]{@{}c@{}}Top $k$\% \\ attn.\end{tabular}}} & \multicolumn{3}{c|}{SNIPS} & \multicolumn{3}{c}{ATIS} \\ \cline{2-7} 
\multicolumn{1}{c|}{}                                                                                & Pos.    & Neg.    & Diff   & Pos.   & Neg.   & Diff   \\ \hline
100\%                                                                                          & \textbf{6.1161}  & 6.1933  & 0.0772 & \textbf{6.3135} & 6.4009 & 0.0874 \\
10\%                                                                                                 & \textbf{3.8986}  & 4.0075  & 0.1089 & \textbf{3.5281} & 3.7878 & 0.2597 \\
5\%                                                                                                  & \textbf{2.8803}  & 2.9955  & 0.1152 & \textbf{2.4589} & 2.7628 & 0.3039 \\ \hline
\end{tabular}
\caption{Average entropy for top $k$\% attention weights for test data. Results confirm attention spikes in positive slot types compared to negative ones. Lower entropy means non-uniform values. Positive slot types (Pos.) are the slot types that appear in an utterance and negative slot types (Neg.) are the ones do not appear.}

\label{tab:entropy_list}
\end{table}

\subsubsection{Qualitative Analysis}
\begin{figure}
    \centering
    \includegraphics[scale=0.45, trim=0.25cm 4.8cm 7.2cm 0cm, clip=true]{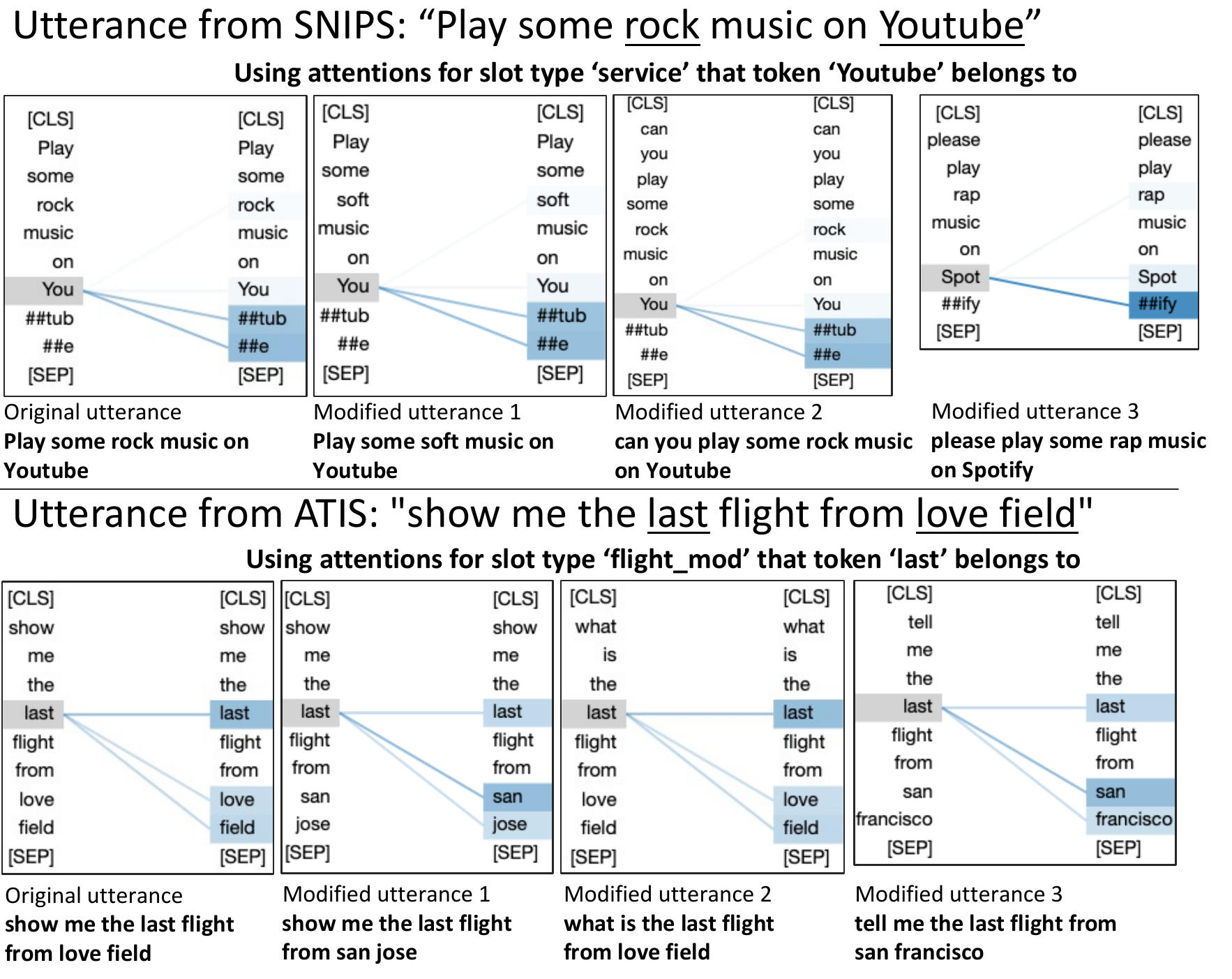}
    \caption{Attention visualizations for modified utterance examples. We can see that the attentions are consistent across modifications, hence, trustworthy.}
    \label{fig_visualize_qualitative}
\end{figure}
Our goal is to inspect whether the model attentions are consistent, trustworthy, and not arbitrary. We performed a small-scale human evaluation to check whether slot type attentions remain similar if the utterances are slightly modified. We randomly selected 20 utterances each, from ATIS and SNIPS. Three types of modifications were applied to each of these utterances: (i) modifying slot values only, (ii) utterance text only, and (iii) both utterance text and slot values. We obtained a total of 6 modifications for each utterance (2 per modification type). We make sure that utterance semantics (original slots types) and the meaning of the utterances do not change due to these modification. Figure~\ref{fig_visualize_qualitative} shows few examples of such modifications. We collected 120 modifications (20 x 6) for each dataset and 240 (120 x 2) in total for both datasets. 

For each original utterance, a positive slot type was selected at random and the corresponding computed attention patterns were compared with the original utterance for each of the modifications. We used a pool of two judges and they were asked to rate the similarity of the attention patterns on a 3-point Likert scale (3 for exactly same, 2 for similarity with some overlap, 1 for no overlap). We obtained 344/360 for ATIS and 354/360 for SNIPS (the total possible score per dataset is 360=120x3). These scores indicate that the attentions for slot types were consistent across the modifications and they are not arbitrary or opaque. This shows that our slot type specific attention mechanism is able to successfully learn patterns that are specific to slot types `explicitly'. These patterns are also consistent, and hence, these slot type specific attention weights can provide a means for trustworthy explanations. See Appendix \ref{sec:appendix} for few more examples.

\section{Conclusion}
We presented a novel joint NLU approach where we jointly model intent detection and slot filling with per slot type attention learning. Slot type specific attention weights enabled the model to provide fine-grained slot explanations and additional slot type features computed through the auxiliary network (i.e., logits) improved model accuracy. To the best of our knowledge, our proposed approach is the first to inherently provide explanations for slot filling while jointly detecting both intents and slots in utterances, without any post-hoc processing. The added transparency from our explainable model is beneficial for further debugging and improvement of models, and for increasing end user trust in personal voice assistants and task oriented dialog systems. In future, we plan to investigate the use of this type of inherently explainable model through auxiliary network constraint enforcement for problems such as text classification and NER.

\section*{Limitations}
Currently, our proposed model provides explanations for the slot filling task, which is challenging because we need to provide explanations for each word in an utterance. It is possible to extend our model to explain intent detection as well, where explanations are given for the classification decision over the entire utterance. We have not yet explored this aspect of explainable intent detection in the joint model. Our approach consists of additional auxiliary network and it needs additional parameters on top of BERT-Base model to properly learn new features and attention weights. For example, for SNIPS dataset, our model uses 120 million trainable parameters (model size is 461MB) compared to 110 million parameters in BERT-Base model (model size is 438MB). Hence, our model is larger than BERT base and the increase in model size will depend on the number of slot types in the dataset. However, in real life, for joint NLU models, the number of slot types is in the order of 10s (less than 100 in most cases) and since we are not considering this type of a model for on-device deployment, we believe this is manageable. Note that number of slot types is around half the size of BIO slot labels.

\bibliography{anthology,custom}
\bibliographystyle{acl_natbib}

\appendix

\section{Appendix}
\label{sec:appendix}

\subsection{Slot type specific attentions visualizations}
\begin{figure*}
    \centering
    \includegraphics[scale=0.65, trim=0cm 1cm 1cm 0cm, clip=true]{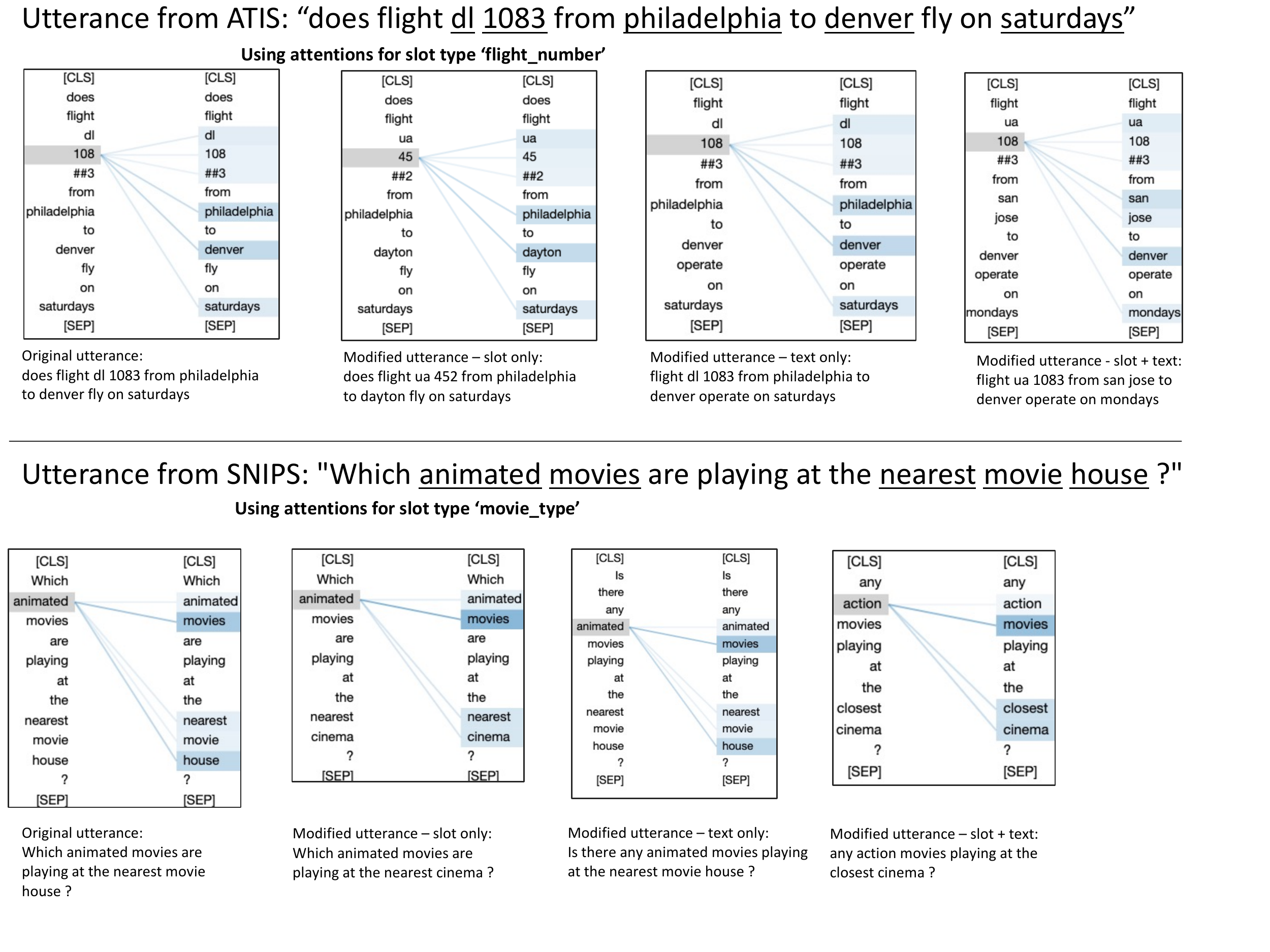}
    \caption{Analyzing attentions for slightly modified utterances where semantics of the modified are the same as originals. Examples show that attentions computed for specific slot types remain very similar.}
    \label{fig_visualize_qualitative_more}
\end{figure*}

Here we show two more utterances taken from ATIS and SNIPS and show visualizations for modifications of the original utterances in Figure~\ref{fig_visualize_qualitative_more}. 

\subsection{Example utterances in qualitative analysis}
Here we show six modifications for the above two randomly selected utterances. The modifications are shown in Table~\ref{tab:example updated utterances}.
\begin{table*}
\small
\begin{tabular}{p{0.75cm}p{4.12cm}p{1.5cm}p{8cm}}
\hline 
\textbf{Dataset} & \textbf{Original Utterance}                                               & \textbf{Modification Type}            & \textbf{Modified Utterance}                                               \\ \hline 
ATIS    & does flight dl 1083 from philadelphia to denver fly  on saturdays  & \multirow{2}{*}{Slot only}   & (1) does flight ua 1083 from philadelphia to denver fly on saturdays \\
        &                                                                &                              & (2) does flight ua 452 from philadelphia to dayton fly on saturdays  \\
        \cline{3-4} 
        &                                                                  & \multirow{2}{*}{Text only}   & (3) flight dl 1083 from philadelphia to denver fly on saturdays      \\ 
        &                                                                  &                              & (4) flight dl 1083 from philadelphia to denver operate on saturdays  \\
        \cline{3-4} 
        &                                                                  & \multirow{2}{*}{Slot + Text} & (5) flight ua 1083 from philadelphia to denver operate on saturdays  \\ 
        &                                                                  &                              & (6) flight ua 1083 from san jose to denver operate on mondays        \\ \hline 
 SNIPS   & Which animated movies are playing at the nearest movie house ?     & \multirow{2}{*}{Slot only}   & (1) Which animated movies are playing at the nearest cinema ?         \\
        &                                                                  &                              & (2) Which action movies are playing at the farthest cinema ?          \\
        \cline{3-4} 
        &                                                                  & \multirow{2}{*}{Text only}   & (3) What animated movies are playing at the nearest movie house ?     \\ 
        &                                                                  &                              & (4) Is there any animated movies playing at the nearest movie house ? \\
        \cline{3-4} 
        &                                                                  & \multirow{2}{*}{Slot + Text} & (5) Is there any action movies playing at the nearest cinema ?        \\ 
        &                                                                  &                              & (6) any action movies playing at the closest cinema ?                \\  \hline                                                      
\end{tabular}
\caption{Showing two original utterances from ATIS and SNIPS and how each utterance gets modified into 6 unique utterances to test slot type attention weights for consistency.}
\label{tab:example updated utterances}
\end{table*}


\subsection{Experiment details}
\label{sec:appendix_parameters}

We used Tesla V100-SXM 32GB gpu configured in a gpu cluster and used CUDA 10.1. Our model for ATIS has 130 million parameters and for SNIPS, it has 120 million parameters. ATIS model has little more parameters because it has more slot types. Saved model size for ATIS dataset is 500MB whereas for SNIPS, it is 461MB.

Best performing development set accuracy values are as follows. For ATIS, intent accuracy and slot F1 values are 98.60 and 98.33, respectively. For SNIPS, they are 99.14 and 97.60, respectively. On average, training takes around 20 - 25 minutes on the above GPU.

\end{document}